\documentclass{article}
\PassOptionsToPackage{sort,square,comma,numbers}{natbib}
\usepackage[final]{nips_2016}

\usepackage[utf8]{inputenc} % allow utf-8 input
\usepackage[T1]{fontenc}    % use 8-bit T1 fonts
\usepackage{hyperref}       % hyperlinks
\usepackage{url}            % simple URL typesetting
\usepackage{booktabs}       % professional-quality tables
\usepackage{amsfonts}       % blackboard math symbols
\usepackage{nicefrac}       % compact symbols for 1/2, etc.
\usepackage{microtype}      % microtypography
\usepackage{graphicx}
\usepackage{caption,subcaption}

\usepackage{algpseudocode}
\usepackage{algorithm}
\usepackage{listings,xcolor}
\usepackage[lighttt]{lmodern}
\usepackage{tikz}
\usepackage{placeins}
\usetikzlibrary{fit,shapes}

\usepackage[compact]{titlesec}
\titlespacing\section{0pt}{4pt plus 2pt minus 2pt}{2pt plus 2pt minus 2pt}
\titlespacing\subsection{0pt}{4pt plus 2pt minus 2pt}{2pt plus 2pt minus 2pt}
\usepackage{booktabs}

\title{Learning to superoptimize programs}

\author{Rudy Bunel, Alban Desmaison, M. Pawan Kumar \& Philip H.S. Torr \\
  Department of Engineering Science\\
  University of Oxford \\
  Oxford, UK\\
  \texttt{\{rudy,alban,pawan\}@robots.ox.ac.uk}, \texttt{philip.torr@eng.ox.ac.uk}\\
  \And
  Pushmeet Kohli\\
  Microsoft Research\\
  Redmond, WA 98052, USA \\
  \texttt{pkohli@microsoft.com}\\
}

\begin{document}

\maketitle

\begin{abstract}
  Superoptimization requires the estimation of the best program for a given
  computational task. In order to deal with large programs, superoptimization
  techniques perform a stochastic search. This involves proposing a modification
  of the current program, which is accepted or rejected based on the improvement
  achieved. The state of the art method uses uniform proposal distributions,
  which fails to exploit the problem structure to the fullest. To alleviate this
  deficiency, we learn a proposal distribution over possible modifications using
  Reinforcement Learning. We provide convincing results on the superoptimization
  of ``Hacker's Delight'' programs.
\end{abstract}

\section{Introduction}
Superoptimization requires us to obtain the optimal program for a computational
task. While modern compilers implement a large set of rewrite rules, they fail
to offer any guarantee of optimality. An alternative approach is to search over
the space of all possible programs that are equivalent to the compiler output,
and select the one that is the most efficient. If the search is carried out in a
brute-force manner, we are guaranteed to achieve superoptimization. However,
this approach quickly becomes computationally infeasible as the number of
instructions and the length of the program grows.

To address this issue, recent approaches have started to use
a stochastic search procedure, inspired by Markov Chain Monte Carlo
sampling~\citep{schkufza2013stochastic}. One of the main factors that governs
the efficiency of this stochastic search is the choice of a proposal
distribution. Surprisingly, the state of the art method,
Stoke~\citep{schkufza2013stochastic} relies on uniform distributions for each of
its components. We argue that this choice fails to fully exploit the power of
stochastic search.

To alleviate the aforementioned deficiency of Stoke, we build a reinforcement
learning framework to estimate a more suitable proposal distribution for the
task at hand. The quality of the distribution is measured as the expected
quality of the program obtained via stochastic search. Using training data,
which consists of a set of input programs, the parameters are learnt via the
REINFORCE algorithm~\citep{williams1992simple}. We demonstrate the efficacy of
our approach on a set of ``Hacker's Delight''~\citep{warren2002hacker} programs.
Preliminary results indicate that a learnt proposal distribution outperforms the
uniform one on novel tasks that were previously unseen during training.
\section{Related Works}

The earliest approached for superoptimization relied on brute-force search. By
sequentially enumerating all programs in increasing length
orders~\citep{granlund1992eliminating,massalin1987superoptimizer}, the shortest
program meeting the specification is guaranteed to be found. As expected, this
approach scales poorly to longer programs or to large instruction sets.
The longest reported synthesized program was 12 instructions long,
on a restricted instruction set~\citep{massalin1987superoptimizer}.

Trading off completeness for efficiency, stochastic
methods~\citep{schkufza2013stochastic} reduced the number of programs to test by
guiding the exploration of the space, using the observed quality of programs
encountered as hints. However, using a generic, unspecific exploratory policy
made the optimization blind to the problem at hand. We propose to tackle this
problem by learning the proposal distribution.

Similar work was done to discover efficient implementation of computation of
value of degree $k$ polynomials~\citep{efficient-math}. Programs were generated
from a grammar, using a learned policy to prioritize exploration. This
particular approach of guided search looks promising to us, and is in spirit
similar to our proposal, although applied on a very restricted case.

Another approach to guide the exploration of the space of programs was to make
use of the gradients of differentiable relaxation of programs. \citet{anc}
attempted this by simulating program execution using recurrent Neural Networks.
This however provided no guarantee that the optimum found was going to correspond
to a real program. Additionally, this method only had the possibility of
performing very local moves, limiting the kind of discoverable transformations.

%\subsection{Learning to optimize}
Outside of program optimization, applying learning algorithms to improve
optimization procedures, either in terms of results achieved or time taken, is a
well studied subject. \citet{doppa2014hc} proposed methods to deal with
structured output spaces, in a ``Learning to search'' framework. However, these
approaches based on Imitation Learning are not directly applicable as we have
access to a valid cost function, and therefore don't need to learn how to
approximate it. More relevant is the recent literature on learning to optimize.
\citet{li2016learning} and \citet{andrychowicz2016learninggd} learns how to
improve on first-order gradient descent algorithms, making use of neural
networks. Our work is similar, as we aim to improve the optimization process. We
differ in that our initial algorithm is a MCMC sampler, on a discrete space, as
opposed to gradient descent on a continuous, unconstrained space.

The training of a Neural Network to generate a proposal distribution to be used
in sequential Monte-Carlo was also proposed by \citet{paige2016inference} as a
way to accelerate inference in graphical models. Additionally, similar
approaches were successfully employed in computer vision problems where data
driven proposals allowed to make inference
feasible~\citep{jampani2015informed,kulkarni2015picture,datadrivenmcmc}.

\newcommand{\mT}{\mathcal{T}}
\newcommand{\mR}{\mathcal{R}}
\newcommand{\cost}[2]{\textnormal{cost}\left(\mathcal{#1}_{#2},\mT\right)}
\newcommand{\move}{\mR \rightarrow \mR^{\star}}
\section{Learning Stochastic Superoptimization}
Stoke performs black-box optimization of a cost function on the space of
programs, represented as a series of instructions. Each instruction is composed
of an opcode, specifying what to execute, and some operands, specifying the
corresponding registers. Each given input program $\mT$ defines a cost function. For a candidate
program $\mR$ called a rewrite, the associated cost is given by:
\begin{equation}
  \label{eq:cost-fun}
  \cost{R}{} = \omega_e \times \textnormal{eq}(\mR, \mT) + \omega_p \times \textnormal{perf}(\mR)
\end{equation}
The term $\textnormal{eq}(\mathcal{R};\mathcal{T})$ measures how well do the outputs of
the rewrite match with the outputs of the reference program when executed. This can be
obtained either by running a symbolic validator or by running test cases, and
accepting partial definition of correctness.
The other term, $\textnormal{perf}(\mathcal{R})$ is a measure of the execution time of the
program. An approximation can be the sum of the latency of all the instructions
in the program. Alternatively, timing the program on some test cases can be used.

To find the optimum of this cost function, Stoke runs an MCMC sampler, using the
Metropolis algorithm. This allows to sample from the probability
distribution induced by the the cost function:

\begin{equation}
  \label{eq:proba-target}
  p(\mathcal{R}; \mathcal{T}) = \frac{1}{Z} \exp( - \cost{R}{}) ),
\end{equation}
where $\mathcal{R}$ is the proposed rewrite, $\mathcal{T}$ is the input
program.

The sampling is done by proposing random moves $\move$, sampled from a proposal
distribution $q(\mathcal{R^{\star}} | \mathcal{R})$. An acceptance criterion is
computed, and used as the parameter of a Bernoulli distribution, to decide
whether or not the move is accepted.
\begin{equation}
  \label{eq:sym-acceptance}
  \alpha(\move, \mathcal{T}) = \min \left(1 , \frac{p(\mathcal{R^{\star}; \mathcal{T}})}{p(\mathcal{R; \mathcal{T}})}\right).
\end{equation}

The proposal distribution $q$ originally used in \citep{schkufza2013stochastic}
is a hierarchical model, whose detailed structure distribution can be found in
Appendix~\ref{sec:genmodel}. Uniform distributions were used for each of the
elementary probability distributions the model sample from. This corresponds to
a specific instantiation of the general approach. We propose to learn those
probability distribution so as to maximize the probability of reaching the best
programs.

The cost function defined in equation~(\ref{eq:cost-fun}) corresponds to what we
want to optimize. Under a fixed computational budget to perform program
superoptimization in less than $T$ iterations, we are interested in having the
lowest possible cost at the end. As different programs have different runtimes
and therefore different associated costs, we need to perform normalization. As
normalized loss function, we use the ratio between the best rewrite found and
the cost of the initial unoptimized program $\mR_0$. Given that our optimization
procedure is stochastic, we will need to consider the expected cost as our loss.
This expected loss is a function of the parameters $\theta$ of our proposal
distribution. The objective function of our ``meta-optimization'' problem is
therefore:

\begin{equation}
  \label{eq:objective}
  \mathcal{L}(\theta) = \mathbb{E}_{\{\mR_t\} \sim q_{\theta}}\left[ \frac{\min_{t=0..T}\cost{R}{t}}{\cost{R}{0}} \right]
\end{equation}

Our chosen parameterization of $q$ is to keep the hierarchical structure of the
original work of~\citet{schkufza2013stochastic}, and parameterize all separate
probability distributions (over the type of move, the opcodes, the operands, and
the lines of the program) independently. In order to learn them, we will make
use of unbiased estimators of the gradient. These can be obtained using the
REINFORCE algorithm~\citep{williams1992simple}. A helpful way to derive them is
to consider the execution traces of the search procedure under the formalism of
stochastic computation graphs~\citep{schulman2015gradient}. The corresponding graph used can
be found in Appendix~\ref{sec:stocha-graph}.

By instrumenting the Stoke system of \citet{schkufza2013stochastic}, we can
collect the execution traces so as to compute gradients over the outputs of the
probability distributions, which can then be back-propagated. In that way, we can
perform Stochastic Gradient Descent (SGD) over our objective
function~\ref{eq:objective}.

\section{Experiments}
We ran our experiments on the Hacker's delight~\citep{warren2002hacker} corpus,
a collection of 25 bit-manipulation programs, used as benchmark in program
synthesis~\cite{gulwani11loopfree,jha2010oracle,schkufza2013stochastic}. A
detailed description of the task is given in Appendix~\ref{sec:hd_tasks}. Some examples include
identifying whether an integer is a power of two from its binary representation,
counting the number of bits turned on in a register or computing the maximum of
two integers.

In order to have a larger corpus than the twenty-five programs initially
obtained, we generate various starting points for each optimization. This is
accomplished by running Stoke with a cost function where $\omega_p = 0$ in
(\ref{eq:cost-fun}), keeping only the correct programs and filtering out
duplicates. This allows us to create a larger dataset.

We divide the Hacker's Delight tasks into two sets. We train on the first set
and only evaluate performance on the second so as to evaluate the generalization
of our learned proposal distribution. We didn't attempt to learn the probability
distribution over the operands and the program position, only learning the ones
over opcodes and type of move to perform.

The probability distribution learned here are simple categorical distribution.
We learn the parameters of each separate distribution jointly, using a Softmax
transformation to enforce that they are proper probability distribution. We
initialize the training with uniform proposal distribution so the first
datapoints on the graph corresponds to the original system of
~\citep{schkufza2013stochastic}.

\begin{figure}[h]
  \centering
  \begin{subfigure}[b]{0.35\linewidth}
    \includegraphics[width=\linewidth]{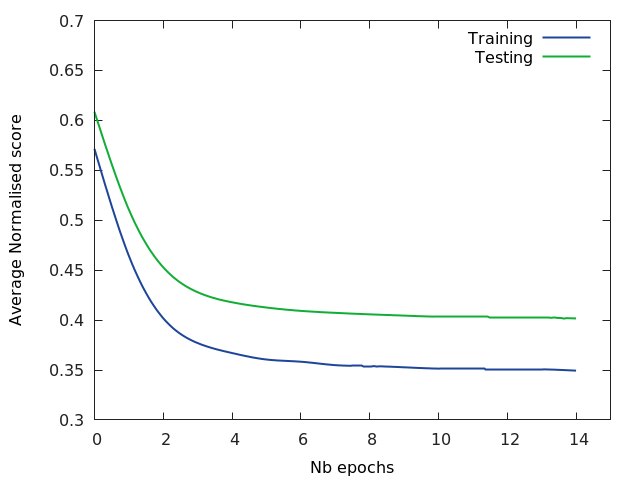}%y
    \caption{\label{fig:bias_train_loss} Evolution of the Objective function.}
  \end{subfigure}%
  \hfill%
  \begin{subfigure}[b]{0.55\linewidth}
    \centering
    \renewcommand{\arraystretch}{1.2}
    \raisebox{55pt}{
    \begin{tabular}{c@{\hspace*{4ex}}cc}
      \toprule
      \textbf{Model} & \textbf{Training}  & \textbf{Test} \\
      \midrule
      Uniform & 57.38\% & 60.90 \% \\
      Learned Bias & 34.93 \% & 40.16\% \\
      \bottomrule
    \end{tabular}%
    }
    \caption{\label{tab:hd_fin}Final improvement score on the Hacker's Delight benchmark.}
  \end{subfigure}
  \caption{\label{tab:hd_comp} Initializing from Uniform
    Distributions as in Stoke~\citep{schkufza2013stochastic}, we manage to
    improve performance by learning the proposal distribution.}
\end{figure}

In our current experiment, the proposal distributions are not conditioned on the
input program. Optimizing them corresponds to finding an ideal proposal
distribution for Stoke. Figure~\ref{fig:bias_train_loss} shows the results. Both the
training and the test loss decreases and it can be observed that the
optimization of program happens faster and that more programs reach the observed
minimum.

\begin{figure}[h]
   \begin{subfigure}{0.3\linewidth}
    \includegraphics[width=0.8\linewidth]{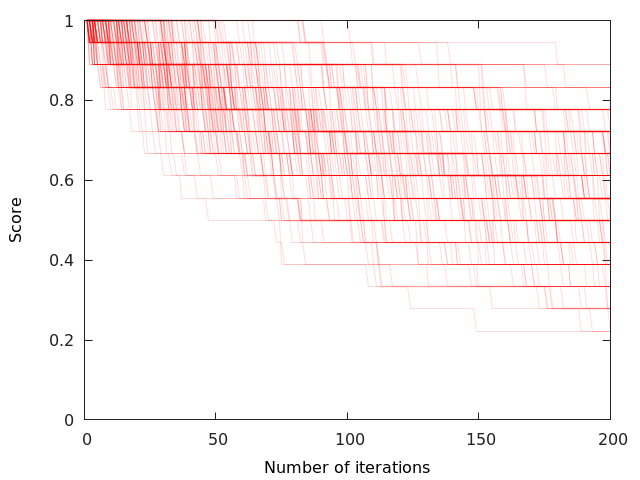}
    \caption{\label{fig:bias_before_train_traces}Optimization Traces}
  \end{subfigure}%
  \hfill%
  \begin{subfigure}{0.3\linewidth}
    \includegraphics[width=0.8\linewidth]{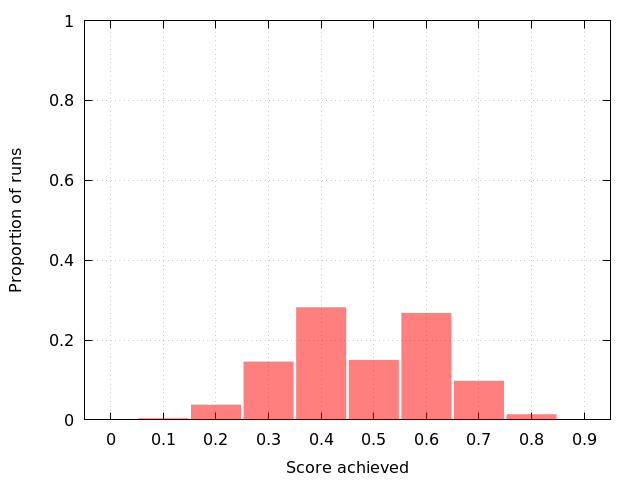}
    \caption{\label{fig:uni_200}Scores after 200 iterations}
  \end{subfigure}
  \hfill%
  \begin{subfigure}{0.3\linewidth}
    \includegraphics[width=0.8\linewidth]{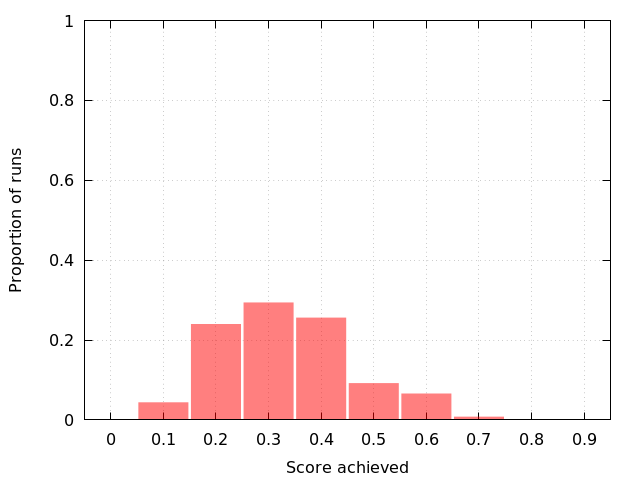}
    \caption{\label{fig:uni_400}Scores after 400 iterations}
  \end{subfigure}
  \caption{Optimization Results using Uniform elementary distributions}
\end{figure}

\begin{figure}[ht]
 \begin{subfigure}{0.3\linewidth}
    \includegraphics[width=0.8\linewidth]{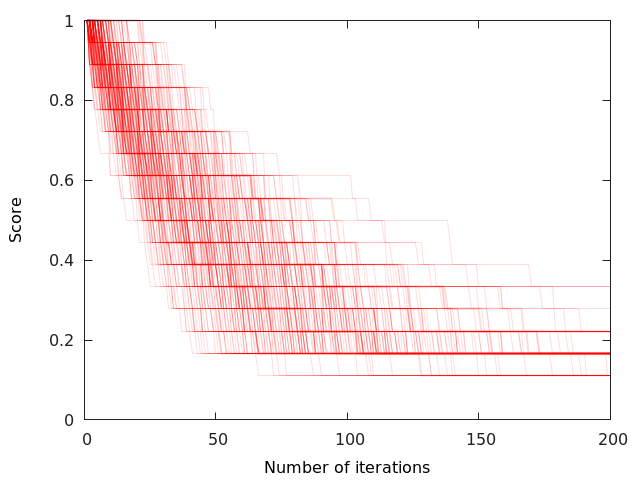}
    \caption{\label{fig:bias_after_train_traces}Optimization Traces}
  \end{subfigure}%
  \hfill%
  \begin{subfigure}{0.3\linewidth}
    \includegraphics[width=0.8\linewidth]{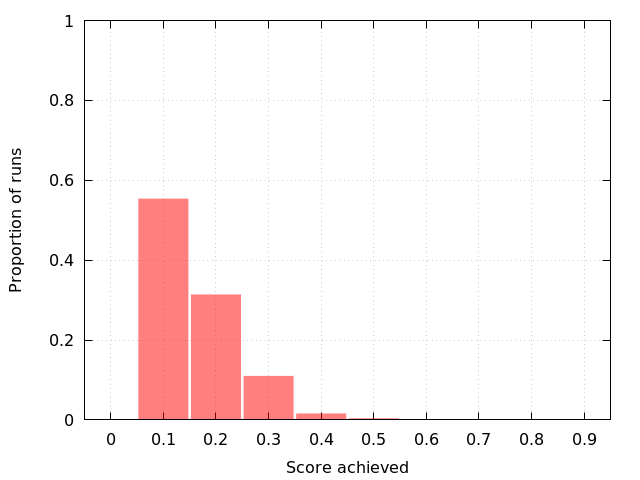}
    \caption{\label{fig:bias_100}Scores after 100 iterations}
  \end{subfigure}
  \hfill%
  \begin{subfigure}{0.3\linewidth}
    \includegraphics[width=0.8\linewidth]{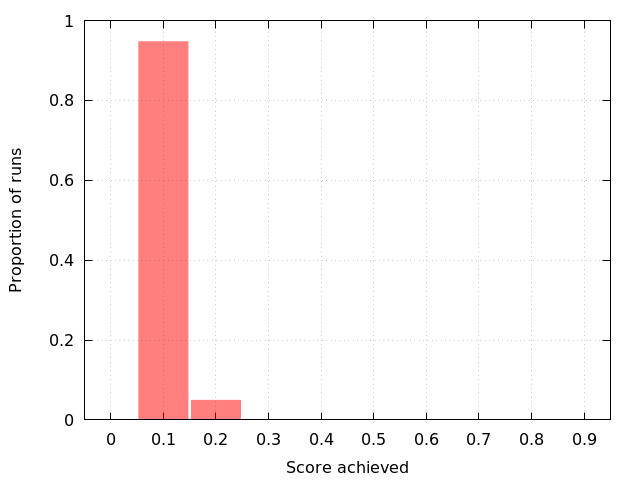}
    \caption{\label{fig:bias_200}Scores after 200 iterations}
  \end{subfigure}
  \caption{Optimization Results using Learned elementary distributions. Even
    with less iterations, better results are achieved than when using Uniform distributions.}
 \end{figure}

\section{Conclusion}
Within this paper, we have shown that learning the proposal distribution of the
stochastic search can lead to significant performance improvement. It is
interesting to compare our approach to the synthesis-style approaches that have
been appearing recently in the Deep Learning community~\citep{graves2014neural}
that aim at learning programs directly using differentiable representations of
programs. We note one advantage that such stochastic search-based approach
yields is that the resulting program can be run independently from the Neural
Network that was used to discover them.

Several improvements are possible to the presented approach. Making the
probability distribution a Neural Network conditioned on the initial input or
on the current state of the rewrite would lead to a more expressive model, while
essentially having similar training complexity. It will however be necessary to
have a richer, more varied dataset to make any evaluation meaningful.

\bibliographystyle{plainnat}
\bibliography{bibliography}

\clearpage
\appendix

\section{Generative model of the program transformations}
\label{sec:genmodel}
In Stoke~\citep{schkufza2013stochastic}, the program transformation are sampled
from a generative model. This process was analysed from the publicly available
code~\cite{stoke-code}.

First, a type of transformation is sampled uniformly from the following proposals method.
\begin{enumerate}
\item\label{stokeMvType:1} \textbf{Add a NOP instruction} Add an empty instruction at a random position in the program.
\item\label{stokeMvType:2} \textbf{Delete an instruction} Remove one of the instruction of the program.
\item\label{stokeMvType:3} \textbf{Instruction Transform} Replace one existing line (instruction + operands) by a new one (New instruction and new operands).
\item\label{stokeMvType:4} \textbf{Opcode Transform} Replace one instruction by another one, keeping the same operands. The new instruction is sampled from the set of compatible instructions.
\item\label{stokeMvType:5} \textbf{Opcode Width Transform} Replace one instruction by another one, with the same memonic. This means that those instructions do the same thing, except that they don't operate on the same part of the registers (for example, will replace \texttt{movq} that move 64-bit of data of the registers by \texttt{movl} that will move 32-bit of data)
\item\label{stokeMvType:6} \textbf{Operand Transform} Replace the operand of a randomly selected instruction by another valid operand for the context, sampled at random.
\item\label{stokeMvType:7} \textbf{Local swap Transform} Swap two instructions in the same ``block''.
\item\label{stokeMvType:8} \textbf{Global Swap transform} Swap any two instructions.
\item\label{stokeMvType:9} \textbf{Rotate transform} Draw two positions in the program, and rotate all the instructions between the two (the last one becomes the first one of the series and all the others get pushed back).
\end{enumerate}

Then, once the type of move has been sampled, the actual move has to be sampled.
To do that, a certain numbers of sampling steps need to happen. Let's take as
example~\ref{stokeMvType:3}.\\

To perform an \textbf{Instruction Transform},
\begin{enumerate}
\item \label{inst-step:1} A line in the existing programs is uniformly chosen.
\item \label{inst-step:2} A new instruction is sampled, from the list of all
  possible instructions.
\item\label{inst-step:3} For each of the arguments of the instruction, sample
  from the acceptable value.
  \item\label{inst-step:4} The chosen line is replaced by the new line that was sampled.
\end{enumerate}

The sampling process of a move is therefore a hierarchy of sampling steps. A
simple way to characterize it is as a generative model over the moves. Depending
on what type of move is sampled, differents series of sampling steps will have
to be performed. For a given move, all the probabilities are sampled independently so
the probability of proposing the move is the product of the probability of
picking each of the sampling steps. The generative model is defined in
Figure~\ref{lst:genCode}. It is going to be parameterized by the the parameters
of each specific probability distribution it samples from. The default Stoke
version uses uniform probabilities over all of those elementary distributions.

\begin{figure}
  \centering
  \lstdefinelanguage{ppl}
  {
    keywords={sample},
    keywordstyle=\color{blue}\bfseries,
    morekeywords=[2]{def},
    keywordstyle=[2]\color{lime}\bfseries,
    morekeywords=[3]{if,return},
    keywordstyle=[3]\bfseries
  }
  \lstset{
    language={ppl},
    basicstyle=\normalfont\ttfamily\footnotesize,
    columns=fixed,
    frame = trbl,
    numbers=left,
    numberstyle=\tiny\ttfamily,
  }
  \newsavebox{\templistingbox}
  \begin{lrbox}{\templistingbox}%
    \begin{minipage}{\textwidth}%
\begin{lstlisting}[language=ppl]
def proposal(current_program):
    move_type = sample(categorical(all_move_type))
    if move_type == 1: % Add empty Instruction
       pos = sample(categorical(all_positions(current_program)))
       return (ADD_NOP, pos)

    if move_type == 2: % Delete an Instruction
       pos = sample(categorical(all_positions(current_program)))
       return (DELETE, pos)

    if move_type == 3: % Instruction Transform
       pos = sample(categorical(all_positions(current_program)))
       instr = sample(categorical(set_of_all_instructions))
       arity = nb_args(instr)
       for i = 1, arity:
           possible_args = possible_arguments(instr, i)
           % get one of the arguments that can be used as i-th
           % argument for the instruction 'instr'.
           operands[i] = sample(categorical(possible_args))
       return (TRANSFORM, pos, instr, operands)

    if move_type == 4: % Opcode Transform
       pos = sample(categorical(all_positions(current_program)))
       args = arguments_at(current_program, pos)
       instr = sample(categorical(possible_instruction(args)))
       % get an instruction compatible with the arguments
       % that are in the program at line pos.
       return(OPCODE_TRANSFORM, pos, instr)

    if move_type == 5: % Opcode Width Transform
       pos = sample(categorical(all_positions(current_program))
       curr_instr = instruction_at(current_program, pos)
       instr = sample(categorical(same_memonic_instr(curr_instr))
       % get one instruction with the same memonic that the
       % instruction 'curr_instr'.
       return (OPCODE_TRANSFORM, pos, instr)

    if move_type == 6: % Operand transform
       pos = sample(categorical(all_positions(current-program))
       curr_instr = instruction_at(current_program, pos)
       arg_to_mod = sample(categorical(args(curr_instr)))
       possible_args = possible_arguments(curr_instr, arg_to_mod)
       new_operand = sample(categorical(possible_args))
       return (OPERAND_TRANSFORM, pos, arg_to_mod, new_operand)

    if move_type == 7: % Local swap transform
       block_idx = sample(categorical(all_blocks(current_program)))
       possible_pos = pos_in_block(current_program, block_idx)
       pos_1 = sample(categorical(possible_pos))
       pos_2 = sample(categorical(possible_pos))
       return (SWAP, pos_1, pos_2)

    if move_type == 8: % Global swap transform
       pos_1 = sample(categorical(all_positions(current_program)))
       pos_2 = sample(categorical(all_positions(current_program)))
       return (SWAP, pos_1, pos_2)

    if move_type == 9: % Rotate transform
       pos_1 = sample(categorical(all_positions(current_program)))
       pos_2 = sample(categorical(all_positions(current_program)))
       return (ROTATE, pos_1, pos_2)
\end{lstlisting}
    \end{minipage}%
  \end{lrbox}%
  \usebox{\templistingbox}
  \caption{\label{lst:genCode}Generative Model of a Transformation}
\end{figure}

The criterion described in equation~(\ref{eq:sym-acceptance}) is justified at the
condition that the proposal distribution is symmetric, that is, $q(\mR^{\star} |
\mR) = q(\mR | \mR^\star)$. In that case, in the limit, the distribution of
states visited by the sampler will be $p$, making the optimal program the most
sampled~\citep{metropolis1953equation}.

By learning the proposal distribution, we won't necessarily maintain the
symmetry property. Even when using only uniform elementary distributions as
in~\citep{schkufza2013stochastic}, the proposal distribution is not symmetric.
An example showing the non-symmetric characteristic is the case of the
\textbf{Instruction Transform} move.

If the proposal is to replace an instruction with two arguments
by one with one argument, the probability of the proposal will be:

\begin{equation}
  q(\mR^\star | \mR) = \frac{1}{n_{\textnormal{\small moves}}} \times \frac{1}{n_{\textnormal{\small opcodes}}} \times \frac{1}{n_\textnormal{\small operands}},
\end{equation}
while the reverse proposal would be:
\begin{equation}
  q(\mR | \mR^\star) = \frac{1}{n_{\textnormal{\small moves}}} \times \frac{1}{n_{\textnormal{\small opcodes}}} \times \frac{1}{n^2_\textnormal{\small operands}},
\end{equation}

As a consequence, the proposal distribution is not symmetric and the properties
of the Metropolis algorithm~\citep{metropolis1953equation} won't apply. Even
without guarantees in the limit, the whole process can still be understood as an
hill-climbing algorithm with a stochastic component to avoid getting stuck in
local maxima.

Another potential solution would be to use the Metropolis-Hastings criterion to
replace the simpler Metropolis criterion~(\ref{eq:sym-acceptance}):
\begin{equation}
  \label{eq:mh-acceptance}
  \alpha(\move, \mathcal{T}) = \min \left(1 , \frac{p(\mathcal{R^{\star}; \mathcal{T}}) q(\mathcal{R}| \mathcal{R^{\star}})}{p(\mathcal{R; \mathcal{T}}) q(\mathcal{R^{\star}}| \mathcal{R})}\right).
\end{equation}
However, this involves developping an inverse model of the proposed moves to
find for each move, the reverse move that would correspond to its undoing and
estimate their probabilities. In the current form of the proposal distribution
in Figure~\ref{lst:genCode}, not all moves have a direct reverse move. For
example, \textbf{Delete} does not.

\clearpage
\FloatBarrier
\section{Metropolis algorithm as a Stochastic Computation Graph}
\label{sec:stocha-graph}
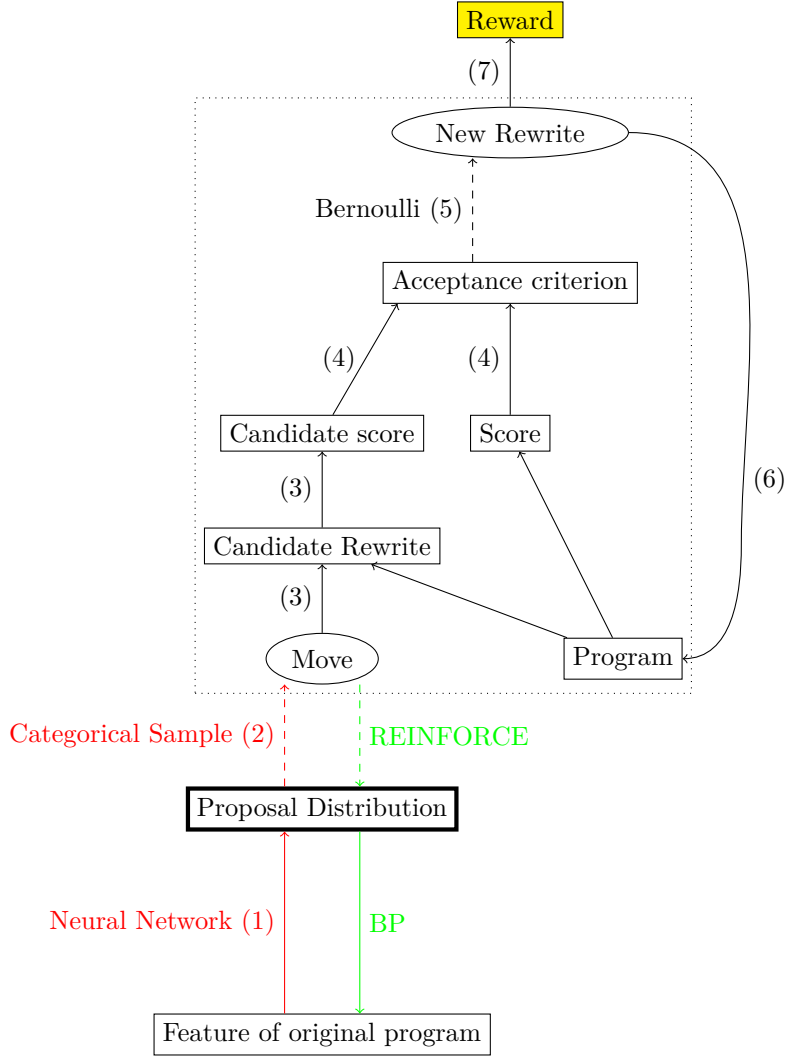
\begin{figure}[h]
  \centering
  \begin{tikzpicture}

  % Network
  \node[draw](feat) at (0,-1) {Feature of original program};
  \node[draw, ultra thick](proposal) at (0, 2) {Proposal Distribution};
  \draw[->, draw=red] ([xshift=-.5cm]feat.north) to node[midway,left] {\color{red}Neural Network (1)} ([xshift=-.5cm]proposal.south);
  \draw[->, draw=green] ([xshift=.5cm]proposal.south) to node[midway,right] {\color{green}BP} ([xshift=.5cm]feat.north);

  % Move
  \node[draw, ellipse](move) at (0,4) {Move};
   \draw[dashed,->, draw=red] ([xshift=-.5cm]proposal.north) to node[ midway,left] {\color{red}Categorical Sample (2)} ([xshift=-.5cm]move.south);
  \draw[dashed, ->, draw=green] ([xshift=.5cm]move.south) to node[midway,right] {\color{green}REINFORCE} ([xshift=.5cm]proposal.north);

  % Program
  \node[draw](prog) at (4,4) {Program};

  % Rewrite
  \node[draw](candidate) at (0,5.5) {Candidate Rewrite};
  \draw[->] (move) to node[midway,left]{(3)} (candidate);
  \draw[->] (prog) to (candidate);

  % Rewrite score
  \node[draw](newscore) at (0,7) {Candidate score};
  \draw[->](candidate) to node[midway,left]{(3)} (newscore);

  % Score
  \node[draw](currscore) at (2.5, 7) {Score};
  \draw[->] (prog) to (currscore);

  % Acceptance Criterion
  \node[draw](acceptance) at (2.5, 9) {Acceptance criterion};
  \draw[->] (newscore) to node[midway,left]{(4)} ([xshift=-1.5cm]acceptance.south);
  \draw[->] (currscore) to node[midway,left]{(4)} (acceptance);

  % Result, accepted or not
  \node[draw, ellipse](rewrite) at (2.5, 11) {New Rewrite};
  \draw[dashed, ->] ([xshift=-.5cm]acceptance.north) to node[midway,left] {Bernoulli (5)} ([xshift=-.5cm]rewrite.south);

  \node[draw, dotted, fit=(move) (prog) (rewrite) (candidate) (newscore) (acceptance)]{};

  \draw[->] (rewrite.east) to[out=0,in=90] node[very near end, right]{(6)} ([xshift=4cm]candidate.east)
                           to[out=-90, in=0] (prog.east);

  \node[draw,fill=yellow](reward) at (2.5, 12.5) {Reward};
  \draw[->] (rewrite) to node[midway,left]{(7)} (reward);

\end{tikzpicture}
  \caption{\em{Stochastic Computation Graph of the Metropolis algorithm used
    for program superoptimization.  Round nodes are stochastic nodes and square
    ones are deterministic. Red arrows corresponds to computation done in
  the forward pass that needs to be learned while green arrows correspond to the
backward pass. Full arrow represent deterministic computation and dashed arrow
represent stochastic ones. The different steps of the forward pass are:\\
(1) Based on features of the reference program, the proposal distribution $q$
  is computed.\\
(2) A random move is sampled from the probability distribution and we keep
  track of the probability of taking this move.\\
(3) The score of the rewrite that would be obtained by applying the chosen
  move is measured experimentally.\\
(4) The acceptance criterion for the move is computed.\\
(5) The move is accepted with a probability equal to the acceptance criterion.\\
(6) Move 2-7 are repeated $N$ times.\\
(7) The reward is observed, corresponding to the best program obtained during
the search.\\
}}
  \label{fig:m-stocha-compg}
\end{figure}

\clearpage
\section{Hacker's delight tasks}
\label{sec:hd_tasks}
The 25 tasks of the Hacker's delight~\cite{warren2002hacker} datasets are the following:

\begin{enumerate}
\item Turn off right-most one bit
\item Test whether an unsigned integer is of the form $2^(n-1)$
\item Isolate the right-most one bit
\item Form a mask that identifies right-most one bit and trailing zeros
\item Right propagate right-most one bit
\item Turn on the right-most zero bit in a word
\item Isolate the right-most zero bit
\item Form a mask that identifies trailing zeros
\item Absolute value function
\item Test if the number of leading zeros of two words are the same
\item Test if the number of leading zeros of a word is strictly less than of
  another work
\item Test if the number of leading zeros of a word is less than of
  another work
\item Sign Function
\item Floor of average of two integers without overflowing
\item Ceil of average of two integers without overflowing
\item Compute max of two integers
\item Turn off the right-most contiguous string of one bits
\item Determine if an integer is a power of two
\item Exchanging two fields of the same integer according to some input
\item Next higher unsigned number with same number of one bits
\item Cycling through 3 values
\item Compute parity
\item Counting number of bits
\item Round up to next highest power of two
\item Compute higher order half of product of x and y
\end{enumerate}

\end{document}